\documentclass{article}
\usepackage{spconf,amsmath,graphicx}

\usepackage{arydshln}
\usepackage{booktabs}

\usepackage{amsmath}
\usepackage{amssymb}
\usepackage{algorithm2e}
\usepackage{subfigure}
\usepackage{graphicx}
\usepackage{multirow}
\usepackage{cite}
\usepackage{color,soul}

\title{ARSM GRADIENT ESTIMATOR FOR SUPERVISED LEARNING TO RANK}

\name{Siamak Zamani Dadaneh$^{\dagger \ast}$ \qquad Shahin Boluki$^{\dagger \ast}$ \qquad Mingyuan Zhou$^{\ddagger}$ \qquad Xiaoning Qian$^{\dagger}$ \thanks{$^{\ast}$ These authors contributed equally to this work}}

			\address{$^{\dagger}$ Texas A\&M University, Department of Electrical and Computer Engineering\\ \quad College Station, Texas, USA \\
			    $^{\ddagger}$ University of Texas at Austin, Department of Information, Risk, and Operations Management\\ \quad Austin, Texas, USA}

\begin{document}
%
\maketitle

\begin{abstract}
We propose a new model for supervised learning to rank. In our model, the relevance labels are assumed to follow a categorical distribution whose probabilities are constructed based on a scoring function. We optimize the training objective with respect to the multivariate categorical variables with an unbiased and low-variance gradient estimator. Learning-to-rank methods can generally be categorized into pointwise, pairwise, and listwise approaches. Although our scoring function is pointwise, the proposed framework permits flexibility over the choice of the loss function. In our new model, the loss function need not be differentiable and can either be pointwise or listwise. Our proposed method achieves better or comparable results on two datasets compared with existing pairwise and listwise methods.
\end{abstract}
\section{Introduction}
\label{sec:intro}

Learning to rank is fundamental to information retrieval, E-commerce, and many other applications, for ranking items~\cite{liu2009learning}. In this work we focus on document retrieval without loss of generality. Document retrieval (i.e., document ranking) has applications in large-scale item search engines, which can generally be described as follows: There is a collection of documents (items). Given a query (e.g. a query entered by a user in the search engine), the ranking function assigns a score to each document, quantifying the relative relevance of the document to the query. The documents are ranked in the descending order based on these scores and the top ranked ones are returned.

Traditional approaches rank documents based on unsupervised models of words appearing in the documents and query and do not need any training \cite{croft2010search}. Supervised machine learning for ranking has become popular due to the availability of more signals related to relevance of documents, such as click items or search log data \cite{liu2009learning}.

The bulk of machine learning methods for learning to rank can roughly be categorized as \emph{pointwise}, \emph{pairwise} and \emph{listwise} methods. Pointwise methods cast the ranking problem as a regression problem for predicting relevance scores \cite{cossock2006subset} or a multiple ordinal classification to predict categorical relevance levels \cite{li2008mcrank}. Pairwise approaches take document pairs as
instances in learning, and formalize the learning-to-rank problem as that of classification. More precisely, they collect document pairs to query the relative ranking from the underlying unknown ranking lists. 
Classification models are then trained with the labeled data for ranking \cite{joachims2002optimizing}. Finally, listwise methods use ranked document lists instead of document pairs as instances in learning and define an optimization loss function over the entire ranked list(s) \cite{cao2007learning}.

In this paper, we propose a new framework for \emph{supervised learning to rank}. Specifically, we define a scoring function that maps the input vector of features for a document to the probability parameters of a categorical distribution, where each category represents the relative relevance of the input document to the query. We then define the objective function of learning-to-rank as the expectation of a loss function, which determines the distance between predicted and true relevance labels of the input document, with respect to the categorical distribution parameterized by the scoring function. To achieve a rich family of ranking algorithms, we employ neural networks as scoring functions.

Due to its novel discrete structure, we exploit stochastic gradient based optimization to learn the parameters of the scoring function. The main difficulty arises when backpropagating the gradients through categorical variables. The recently proposed augment-REINFORCE-swap-merge
(ARSM) \cite{yin2019arsm} gradient estimator provides a natural solution with unbiased low-variance gradient updates during the training of our proposed learning-to-rank framework. ARSM first uses variable augmentation, REINFORCE \cite{williams1992simple}, and Rao-Blackwellization \cite{casella1996rao} to re-express the gradient as an
expectation under the Dirichlet distribution, then uses variable swapping to construct differently
expressed but equivalent expectations, and finally shares common random numbers between these expectations to achieve significant variance reduction.

The proposed framework, hereby referred to as \emph{ARSM-L2R}, has main advantage over the existing learning-to-rank methods. More precisely, due to the utilization of ARSM gradient estimator, the loss function assessing the distance between predicted and true document relevance labels need not be differentiable. This significantly enriches the choices of loss functions that can be employed. Specifically, in our experiments, we optimize the truncated normalized discounted cumulative gain (NDCG) \cite{jarvelin2000ir}.

Comprehensive experiments conducted on benchmark datasets demonstrate that our proposed ARSM-L2R method achieves better or comparable results with pairwise and listwise approaches in terms of common ranking metrics such as truncated NDCG and mean average precision (MAP).

The remainder of this paper is organized as follows. In Section~2, we present the methodology, including the new formulation of ARSM-L2R for supervised learning to rank, and its parameter estimation using Monte Carlo gradient estimates. Section~3 provides comprehensive experimental results for comparison with several existing learning-to-rank methods. The paper is concluded in Section~4.

\section{ARSM-L2R}
\label{sec:met}

\subsection{Supervised learning to rank}

In the supervised learning-to-rank setting, a set of queries $Q=\{q^{(1)},...,q^{(N)}\}$ is given. Each query $q^{(i)}$ is associated with a list of documents $\boldsymbol{d}^{(i)} = [d_1^{(i)},...,d_{n^{(i)}}^{(i)}]$, where $d_j^{(i)}$ and $n^{(i)}$ denote the $j$th document and size of $\boldsymbol{d}^{(i)}$ respectively. In addition, a list of scores $\boldsymbol{y}^{(i)} = [y_1^{(i)},...,y_{n^{(i)}}^{(i)}]$ is available for each list of documents $\boldsymbol{d}^{(i)}$. The score $y_j^{(i)}$ represents the relevance degree of document $d_j^{(i)}$ to query $q^{(i)}$, and can be a judgment score explicitly or implicitly given by humans \cite{cao2007learning}. Higher scores imply more relevant documents.

For each query-document pair $(q^{(i)},d_j^{(i)})$, a $P$-dimensional vector of features $\boldsymbol{x}_j^{(i)}$ is constructed. The training set is represented as $\big\{ (\boldsymbol{x}^{(i)},\boldsymbol{y}^{(i)}) \big\}_{i=1}^{N}$. The objective of learning is to create ranking functions that map the input query-document features to scores resembling the true relevant scores. In the following discussions, we drop the query index $(i)$ to avoid cluttering the notations.

In this paper, we formulate the supervised learning-to-rank problem as maximizing an objective, expressed as an expectation over multivariate categorical variables. More specifically, given $n$ documents for a query, let $z_j \in \{1,\ldots,C\}$ denote the relevance label for $j$th document, where $C$ is the number of possible levels of relevance for each document. In our proposed generative model, each $z_j$ is distributed according to a categorical distribution whose probabilities are constructed based on a scoring function $\mathcal{T}_{\boldsymbol{\theta}}:\mathbb{R}^P \rightarrow \mathbb{R}^C$ parameterized by $\boldsymbol{\theta}$:
\begin{equation}
    z_j \sim \mbox{Cat}(\sigma(\boldsymbol{\phi}_j)), \quad \boldsymbol{\phi}_j = \mathcal{T}_{\boldsymbol{\theta}}(\boldsymbol{x}_j).
\end{equation}
Here $\sigma(\boldsymbol{\phi}_j) = (e^{\phi_{j1}},...,e^{\phi_{jC}})/\sum_{c=1}^{C} e^{\phi_{jc}}$ is the softmax function. We use multi-layer perceptrons (MLPs) as scoring functions, thus $\boldsymbol{\theta}$ corresponds to the collection of weight matrices of MLPs. For each realization of categorical variables $\boldsymbol{z}=(z_1,...,z_n)$, we employ a loss function $\ell$ to determine their distance from the true labels $\boldsymbol{y}=(y_1,...,y_n)$. We then define the learning-to-rank optimization problem as finding:
\begin{eqnarray}
    \hat{\boldsymbol{\theta}} &=& \arg \min_{\boldsymbol{\theta}} \mathbb{E}_{z \sim \prod_{j=1}^n \mbox{Cat}(z_j;\sigma(\boldsymbol{\phi}))}[\ell(\boldsymbol{z},\boldsymbol{y})] \nonumber\\ &:=& \arg \min_{\boldsymbol{\theta}} \mathcal{E}(\boldsymbol{\Phi}),
    \label{eq:obj}
\end{eqnarray}
where $\ell(\cdot,\cdot)$ can be any loss function measuring the dissimilarity of two vectors of ordinal labels. We resort to stochastic gradient based methods to solve the optimization problem in~(\ref{eq:obj}). Backpropagating the gradient through discrete latent variables have been recently studied extensively \cite{tucker2017rebar,yin2019arsm,grathwohl2017backpropagation,boluki2020lbd}. For optimizing (\ref{eq:obj}), the challenge lies in developing a low-variance and unbiased estimator for its gradient with respect to $\boldsymbol{\phi}$, which is denoted by $\nabla_{\boldsymbol{\phi}} \mathcal{E}(\boldsymbol{\Phi})$. 

\subsection{ARSM gradient estimator}

We employ Augment-REINFORCE-Swap-Merge (ARSM) gradient estimator for training the scoring functions described in the previous section. To describe this algorithm, we start by the simple objective function $\mathcal{E}(\boldsymbol{\phi}):=\mathbb{E}_{z \sim \mbox{Cat}(\sigma(\boldsymbol{\phi}))}[f(z)]$ with respect to a univariate categorical variable, where $f(z)$ is the reward function and $\boldsymbol{\phi}:=(\phi_1,\ldots,\phi_C)$. In the \emph{augmentation} step, the gradient of $\mathcal{E}(\boldsymbol{\phi})$ can be expressed as an expectation under a Dirichlet distribution as
\begin{eqnarray}
\nabla_{\phi_c} \mathcal{E}(\boldsymbol{\phi}) &=& \mathbb{E}_{\boldsymbol{\pi} \sim \mbox{Dir}(\boldsymbol{1}_C)}[f(z)(1-C\pi_c)], \nonumber\\
z &:=& \arg \min_{k \in \{1,\ldots,C\}} \pi_k e^{-\phi_k}.
\label{eq:ar}
\end{eqnarray}
Given the vector $\boldsymbol{\pi}$, we denote the vector obtained after swapping $k$th and $m$th elements of $\boldsymbol{\pi}$ as $\boldsymbol{\pi}^{m \rightleftharpoons k}:=(\pi_1^{m \rightleftharpoons k},\ldots,\pi_C^{m \rightleftharpoons k})$, where $\pi_m^{m \rightleftharpoons k}=\pi_k$, $\pi_k^{m \rightleftharpoons k}=\pi_m$ and for $c \notin \{m,k\}$ we have $\pi_c^{m \rightleftharpoons k}=\pi_c$. Exploiting the symmetrical property $\boldsymbol{\pi}^{m \rightleftharpoons k} \sim \mbox{Dir}(\boldsymbol{1}_C)$, and sharing common random numbers between different expectations to significantly reduce
Monte Carlo integration variance leads to another unbiased estimator referred as ARS estimator:
\begin{eqnarray}
\nabla_{\phi_c} \mathcal{E}(\boldsymbol{\phi}) &=& \mathbb{E}_{\boldsymbol{\pi} \sim \mbox{Dir}(\boldsymbol{1}_C)}[f_{\Delta}^{c \rightleftharpoons k}(1-C\pi_k)], \nonumber\\
f_{\Delta}^{c \rightleftharpoons k} &:=& f(z^{c \rightleftharpoons k}) - \frac{1}{C}\sum_{m=1}^{C} f(z^{m \rightleftharpoons k}),
\label{eq:ars}
\end{eqnarray}
where $z^{c \rightleftharpoons k} := \arg \min_{k' \in \{1,\ldots,C\}} \pi_{k'}^{c \rightleftharpoons k} e^{-\phi_{k'}}$ and $k$ is the reference category. Finally, the ARS estimator can be further improved by considering all swap operations, and adding a \emph{merge} step to construct the ARSM estimator as
\begin{eqnarray}
\nabla_{\phi_c} \mathcal{E}(\boldsymbol{\phi}) = \mathbb{E}_{\boldsymbol{\pi} \sim \mbox{Dir}(\boldsymbol{1}_C)} \Big[\sum_{k=1}^C f_{\Delta}^{c \rightleftharpoons k}(1/C-\pi_k) \Big].
\label{eq:arsm}
\end{eqnarray}

\subsection{ARSM for learning to rank}
To employ ARSM for learning to rank, we need to consider the optimization problem with respect to the multivariate categorical variables $\boldsymbol{z}=(z_1,...,z_n)$. Let $\boldsymbol{z}^{c \rightleftharpoons k}=(z_1^{c \rightleftharpoons k},...,z_n^{c \rightleftharpoons k})$ denote the multivariate swapping whose elements are defined, similar to those in (\ref{eq:ars}) and (\ref{eq:arsm}), as $z_j^{c \rightleftharpoons k} := \arg \min_{k' \in \{1,...,C\}} \pi_{jk'}^{c \rightleftharpoons k} e^{-\phi_{jk'}}$. Then the multivariate extension of ARSM gradient estimator for the learning-to-rank objective in (\ref{eq:obj}) can be expressed as \cite{yin2019arsm}:
\begin{equation}
    \nabla_{\phi_{jc}} \mathcal{E}(\boldsymbol{\Phi}) = \mathbb{E}_{\boldsymbol{\Pi} \sim \prod_{j=1}^n \mbox{Dir}(\boldsymbol{\pi}_j;\boldsymbol{1}_C)} \Big[\sum_{k=1}^C \ell_{\Delta}^{c \rightleftharpoons k}(1/C-\pi_{jk}) \Big],
    \label{eq:arsm_l2r}
\end{equation}
where $\ell_{\Delta}^{c \rightleftharpoons k} = \ell(\boldsymbol{z}^{c \rightleftharpoons k},\boldsymbol{y}) - \frac{1}{C}\sum_{m=1}^{C} \ell(\boldsymbol{z}^{m \rightleftharpoons k},\boldsymbol{y})$. Since we define the categorical distribution parameter $\boldsymbol{\Phi}$ in terms of a neural network with parameters $\boldsymbol{\theta}$, the final gradients are computed using the chain rule as
\begin{eqnarray}
\nabla_{\boldsymbol{\theta}} \mathcal{E}(\boldsymbol{\Phi}) &=& \sum_{j=1}^n \sum_{c=1}^C \nabla_{\phi_{jc}} \mathcal{E}(\boldsymbol{\Phi}) \frac{\partial \phi_{jc}}{\partial \boldsymbol{\theta}} \nonumber\\
&=& \nabla_{\boldsymbol{\theta}} \Big(  \sum_{j=1}^n \sum_{c=1}^C \nabla_{\phi_{jc}} \mathcal{E}(\boldsymbol{\Phi}) \phi_{jc} \Big).
\end{eqnarray}

The estimated gradients are then utilized in a stochastic optimization process to learn the model parameters. Algorithm~1 summarizes the parameter learning for ARSM-L2R.

\subsection{Loss function and rank prediction}
The loss function $\ell(\boldsymbol{z},\boldsymbol{y})$ in (\ref{eq:obj}) measures the dissimilarity between predicted categorical labels $\boldsymbol{z}$ and the true labels $\boldsymbol{y}$. In this work, we utilize the negative truncated NDCG as the loss function of ARSM-L2R. The calculation of NDCG only relies on the sorting of the predicted labels $\boldsymbol{z}$, and the true labels $\boldsymbol{y}$. Furthermore, our experiments show that setting the number of possible levels of relevance $C$ to be higher than the number of true levels in $\boldsymbol{y}$ improves the performance of ARSM-L2R. Hence, for all experiments in this paper we set $C=20$.%

After the parameters of the scoring function are learned in the training phase, the probability of different levels of relevance for the test documents can be calculated by simply passing the documents features through the scoring function. We then construct the final scores of the test documents by a weighted combination of these probabilities, and sort the documents based on these scores. More precisely, given the probability of different labels $p_c$ for a test document, we calculate its overall ranking score as $\sum_{c=1}^C c \times p_c$, where $c \in \{1,2,\ldots,C\}$ and higher values of $c$ correspond to more relevant levels. Our experiments show that the performance of ARSM-L2R is not sensitive to the choice of the weight combination scheme.

\begin{algorithm}[t]
\SetAlgoLined
\SetKwInOut{Input}{input}
\SetKwInOut{Output}{output}
\Input{ Document labels $\boldsymbol{y}$ and query-document features $\boldsymbol{x}$}
\Output{ Parameters $\boldsymbol{\theta}$ of scoring function}
Initialize $\boldsymbol{\theta}$ randomly\;
\While{not converged}{
 Calculate categorical distribution parameters $\boldsymbol{\Phi}=(\boldsymbol{\phi}_1,...,\boldsymbol{\phi}_n) \in \mathbb{R}^{C \times n}$\;
 Sample $\boldsymbol{\pi_j} \sim \mbox{Dirichlet}(\boldsymbol{1}_C)$ for $j=1,...,n$\;
 Let $z_j = \arg\min_{k \in \{1,...,C\}} (\ln \pi_{jk} - \phi_{jk})$ for $j=1,...,n$, to obtain the true categorical labels $\boldsymbol{z}=(z_1,...,z_n)$\;
 Initialize the diagonal of the loss matrix $L \in \mathbb{R}^{C \times C}$ with $\ell(\boldsymbol{z},\boldsymbol{y})$\;
 \For{$(c,k) \in \{(c,k)\}_{c=1:C, k<c}$}{
 Let $z_j^{c \rightleftharpoons k} = \arg\min_{k' \in \{1,...,C\}} (\ln \pi_{jk'}^{c \rightleftharpoons k} - \phi_{jk'})$ for $j=1,...,n$\;
 Denote $\boldsymbol{z}^{c \rightleftharpoons k}=(z_1^{c \rightleftharpoons k},...,z_n^{c \rightleftharpoons k})$\;
 Let $L_{ck} = L_{kc} = \ell(\boldsymbol{z}^{c \rightleftharpoons k}, \boldsymbol{y})$
 }
 Let $\bar{L}_{\cdot k} = \frac{1}{C} \sum_{c=1}^C L_{ck}$ for $k=1,...,C$\;
 Let $g_{\phi_{jc}} = \sum_{k=1}^C (L_{ck} - \bar{L}_{\cdot k})(\frac{1}{C} - \pi_{jk})$ for all $(j,c) \in \{(j,c)\}_{j=1:n, c=1:C}$\;
 Update $\boldsymbol{\theta} = \boldsymbol{\theta} + \eta_{\theta} \nabla_{\boldsymbol{\theta}} \mathcal{E}(\boldsymbol{\Phi})$, with step size $\eta_{\theta}$
 }
 \caption{Parameter inference in ARSM-L2R.}
\end{algorithm}

\section{Experiments}
\label{sec:exp}

\subsection{Datasets}
We evaluate the performance of ARSM-L2R on two widely tested benchmark datasets, including a query set from Million Query track of TREC 2007, denoted as MQ2007 \cite{DBLP:journals/corr/QinL13}, as well as the OHSUMED dataset \cite{qin2010letor}. Each dataset consists of queries, corresponding retrieved documents and labels provided by human experts. The possible relevance labels for each document are ``relevant'', ``partially relevant'', and ``not relevant''. We use the 5-fold partitions provided in the original dataset for 5-fold cross validation in the experiments. In each fold, there are three subsets for
learning: training set, validation set and testing set. The properties of these learning to rank datasets are presented in Table~\ref{tab:data}.

\begin{table}[!h]
  \caption{Properties of learning-to-rank datasets used in the experiments.}
  \vspace{0.03in}
  \label{tab:data}
  \centering
  \begin{tabular}{llll}
    \toprule
    dataset   & \#queries   & \#documents   & \#features\\
    \midrule
    MQ2007  &  1700  & $\sim$25,000,000   & 46   \\
    OHSUMED  & 106 & $\sim$350,000 &  45\\
    \bottomrule
  \end{tabular}\vspace{-0.1in}
\end{table}

\begin{table}[t]
  \caption{Performance of different learning-to-rank methods on MQ2007 dataset.}
  \label{tab:mq7}
  \vspace{0.03in}
  \centering
  \resizebox{\columnwidth}{!}{%
  \begin{tabular}{c|c|c|c|c|c}
    \toprule
    Method   & NDCG@1   & NDCG@3   & NDCG@5   & NDCG@10   & MAP \\
    \midrule
      RankSVM  &  0.4045 & 0.4019 & 0.4072 & 0.4383   &  0.4644 \\
     ListNet  &  0.4002 & 0.4091 & \textbf{0.4170} & \textbf{0.4440}   &  \textbf{0.4652} \\
     AdaRank-MAP  &  0.3821 & 0.3984 & 0.4070 & 0.4335  &  0.4577 \\
     AdaRank-NDCG  &  0.3876 & 0.4044 & 0.4102 & 0.4369   &  0.4602 \\
    \midrule
     ARSM-L2R  &  \textbf{0.4051} & \textbf{0.4112} & 0.4159 & 0.4432  & 0.4608  \\
    \bottomrule

  \end{tabular}%
  }  \vspace{-0.1in}
\end{table}

\begin{table}[t]
  \caption{Performance of different learning-to-rank methods on OHSUMED dataset.}
  \label{tab:oh}
  \vspace{0.03in}
  \centering
  \resizebox{\columnwidth}{!}{%
  \begin{tabular}{c|c|c|c|c|c}
    \toprule
    Method   & NDCG@1   & NDCG@3   & NDCG@5   & NDCG@10   & MAP \\
    \midrule
     RankSVM  &  0.4958 & 0.4207 & 0.4164 & 0.4140  &  0.4468 \\
     ListNet  & 0.5326 & 0.4732 & 0.4432 & 0.4410   &  0.4495 \\
     AdaRank-MAP  &  0.5388 & 0.4682 & 0.4613 & 0.4429  & 0.4418  \\
     AdaRank-NDCG  &  0.5330 & \textbf{0.4790} & \textbf{0.4673} & \textbf{0.4496}  &  0.4424 \\
    \midrule
     ARSM-L2R  & \textbf{0.5601} & 0.4642 & 0.4546 & 0.4460    &  \textbf{0.4503} \\
    \bottomrule
  \end{tabular}%
  }\vspace{-0.1in}
\end{table}

\subsection{Baselines}
We compare the performance of our ARSM-L2R with several state-of-the-art baselines, including a pairwise method of RankSVM \cite{lee2014large}, a listwise method of ListNet \cite{cao2007learning}, and other listwise methods that optimize lower bounds of different evaluation measures: AdaRank-MAP, and AdaRank-NDCG \cite{xu2007adarank}.

\subsection{Evaluation metrics}
We use two popular learning-to-rank scoring functions to compare the predicted rankings of the test documents with their true rankings: truncated Normalized Discounted Cumulative Gain (NDCG@$R$) \cite{jarvelin2000ir} and Mean Average Precision (MAP) \cite{baeza1999modern}.  NDCG (DCG) has the effect of
giving high scores to the ranking lists in which relevant
documents are ranked high. Average Precision (AP) represents the averaged precision over all the positions of documents with relevant label for query $q^{(i)}$. Denoting the ranking list $\boldsymbol{r}^{(i)}$ on $\boldsymbol{d}^{(i)}$, MAP is defined as 
\begin{equation}
    \text{MAP} = \frac{1}{N}\sum_{i=1}^N\text{AP}(q^{(i)})=\frac{1}{N}\sum_{i=1}^N\frac{\sum_{j=1}^{n^{(i)}}w_j^{(i)}y_j^{(i)}}{\sum_{j=1}^{n^{(i)}}y_j^{(i)}},
\end{equation}
where $w_j^{(i)}=\frac{\sum_{l:r_l^{(i)} \leq r_j^{(i)}} y_l^{(i)}}{r_j^{(i)}}$. 
NDCG@$R$ is calculated by
\begin{equation}
\begin{split}
   \text{NDCG@}R &= \frac{1}{N}\sum_{i=1}^N \text{NDCG@}R^{(i)}\\&= \frac{1}{N}\sum_{i=1}^N \frac{1}{\text{IDCG@}R^{(i)}}\sum_{j:r_j^{(i)} \leq R}\frac{2^{y_j^{(i)}}-1}{\text{log}_2(1+r_j^{(i)})},
   \end{split}
\end{equation}
where if $r_{\text{true}}$ represents the true ranking list of $\boldsymbol{d}^{(i)}$, then $\text{IDCG@}R^{(i)}=\sum_{j:r_{\text{true},j}^{(i)} \leq R}\frac{2^{y_j^{(i)}}-1}{\text{log}_2(1+r_j^{(i)})}$. $R$ here represents the truncation level.

\subsection{Implementation details}

For the scoring function neural network, we employ a fully connected neural network with one hidden layer of 500 units and the \emph{tanh} nonlinear activation function. We initialize the weights of the neural network by \emph{Glorot} method \cite{glorot2010understanding}, and train ARSM-L2R using the Adam optimizer \cite{kingma2014adam} with a learning rate of $10^{-4}$. The algorithm is run for a total of 2000 epochs, and the ranking metrics on the validation sets are monitored for choosing the best performing neural network weights. ARSM-L2R is implemented in {\tt Tensorflow} \cite{abadi2016tensorflow}.

\subsection{Results and discussions}

We compare the performance of the different methods based on NDCG@1, NDCG@3, NDCG@5, NDCG@10, and MAP. The results for MQ2007 and OHSUMED datasets are provided in Tables \ref{tab:mq7} and \ref{tab:oh}, respectively. Our ARSM-L2R achieves the highest NDCG@1 and NDCG@3 on the MQ2007 dataset. On the OHSUMED dataset ARSM-L2R has a significantly higher NDCG@1 compared with all the other methods tested. It also shows the best MAP on this dataset. It is worth mentioning that NDCG@1 is one of the most important metrics for ranking systems, since it quantifies the relevance of the top ranked item. It is interesting to note that our method only optimizes a rough approximation of the evaluation metric NDCG, but shows the best performance on both two metrics for each dataset and comparable results for the rest of the metrics on the datasets. Our proposed method achieves better or comparable performance due to utilizing a loss function more directly related to ranking performance and also taking advantage of unbiased and low-variance gradient estimation.

\section{Conclusions}
We have developed a new supervised learning-to-rank model---ARSM-L2R---that generates relevance labels based on a categorical model with probabilities estimated by a MLP. The training objective is optimized with respect to the multivariate categorical variables with an unbiased and low-variance gradient estimator, ARSM. Our method can employ a non-differentiable loss function as opposed to the existing learning-to-rank methods. The experimental results show that ARSM-L2R achieves better or comparable results with pairwise and listwise approaches.


\subsection*{Acknowledgement}
The presented materials are based upon the work supported by the National Science Foundation under Grants CCF-1553281, IIS-1812641, IIS-1812699, and CCF-1934904. We also thank Texas A\&M High Performance Research Computing and Texas Advanced Computing Center for providing computational resources to perform experiments in this work.

\bibliographystyle{IEEEbib}
\bibliography{strings,references}

\end{document}